# Flood Analytics Information System (FAIS)
# Version 4.00 Manual

**Prepared by**
Vidya Samadi

Clemson University, Clemson, South Carolina, USA

September 2021

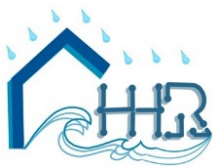
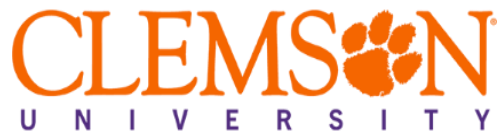
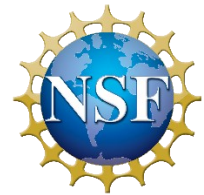

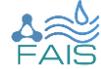

# Contents

**Summary**

This project was the first attempt to use big data analytics approaches and machine learning along with Natural Language Processing (NLP) of tweets for flood risk assessment and decision making. Multiple Python packages were developed and integrated within the Flood Analytics Information System (FAIS). FAIS workflow includes the use of IoTs-APIs and various machine learning approaches for transmitting, processing, and loading big data through which the application gathers information from various data servers and replicates it to a data warehouse (IBM database service). Users are allowed to directly stream and download flood related images/videos from the US Geological Survey (USGS) and Department of Transportation (DOT) and save the data on a local storage. The outcome of the river measurement, imagery, and tabular data is displayed on a web based remote dashboard and the information can be plotted in real-time. FAIS proved to be a robust and user-friendly tool for flood data analysis at regional scale that could help stakeholders for rapid assessment of flood situation and damages. FAIS also provides flood frequency analysis (FFA) to estimate flood quantiles including the associated uncertainties that combine the elements of observational analysis, stochastic probability distribution and design return periods. FFA techniques predict how flow values corresponding to specific return periods or probabilities along a river could change over different design periods. FAIS currently uses multiple probability distributions such as Normal, Lognormal, Gamma, Gumbel, Pearson Type III, Weibull, and Loglogistic distributions to compute FFA for any given flood gauging station in the US. The tool also provides a .CSV report file of the FFA computation that is downloadable from the interface. The use of FFA for any given USGS station provides an easy assessment for the design of engineering structures such as culverts, bridges, and dams. With the many challenges facing existing probability distribution fitting and performance calculation, FFA functions, numerical estimation and uncertainty calculation, and graphical capabilities together with its flexibility to fit multiple distributions, can go a long way. This makes the FAIS application an ideal tool to assess flooding impacts for any USGS gauging station. This project provided support for 4 graduate students and 1 undergraduate student in Civil Engineering and Computer Sciences, and 2 high school students in mathematics and sciences. FAIS is publicly available and deployed on the Clemson-IBM cloud service.

## 1. Introduction

Floods are on the rise globally with the frequent recorded events occurring during the past few years in the US alone. These extreme events pose a considerable threat to human life and results in destructive damage to property, critical infrastructure, and communities (Philips et al., 2018). During flooding events, citizens around the world increasingly act as human sensors and collect and share millions of flood images and videos on social media to record flood magnitude, damage, and impacts. Multimedia images, videos, geotagged texts posted over social media platforms such as Facebook, Twitter, YouTube, Flickr, and other online forums can provide valuable real-time information about flood situation. By using the content and user metadata from volunteered geographic information shared online, we can identify potential at-risk neighborhoods that have been inundated.

In addition, real time surveillance cameras have been installed by several agencies such as the US Geological Survey (USGS) across numerous river networks to meet the need for timely assessment of flood situational awareness (Donratanapat et al., 2020). These real time videos/images can be used to track increasing water levels during a storm and continuously monitor the potential impacts of flooding on nearby locations. Videos and time lapse images can also be processed to extract image frames and related information, which can be used to measure a range of flood characteristics such as flood depth, inundation areas, etc. Indeed, accurate and efficient assessment of real time images is crucially important to assess road and other critical infrastructure conditions during storm. This information provides timely and useful details on the hazards to avoid when flooding has occurred. This includes areas to avoid when using a vehicle, and safety and health hazards such as downed electrical lines, flooded roads, etc.

Developing a pipeline to gather the data and images, and tweets and identify at-risk locations to flooding proved to be useful to assess damage in Sao Paulo- Brazil (de Assis et al, 2016), Jakarta-Indonesia (Eilander et al, 2016), the River Elbe-Germany (Herfort et al, 2014), and across Great Britain (Barker and Macleod, 2018). As needs for flooding impacts assessment and insights grow, stakeholders are facing fragmented data environments and warehouses with multiple technologies—often on multiple web services. There is a need to automate Big data and crowd sourced information collection in real-time and create a map-based dashboard to better determine at-risk locations and flood situations. Indeed, with the new advancement in technologies, there is an opportunity to gather and combine social media data with ground-based observations and imagery and translate this information into a web-based application to monitor and assess flooding hazards and to communicate this information with citizenry in real time. To address these needs, we developed Flood Analytics Information System (FAIS) as a data engineering and analytics pipeline, based on real-time USGS river level information, natural language processing (NLP) of tweets, and river and traffic web cameras imagery.

FAIS allows the user to directly download flood related data from USGS and visualize the data in real time. The outcome of the river measurement, imagery, and tabular data can be also displayed in a web based remote dashboard and the information can be plotted in real-time. Twitter Application Programming Interfaces (APIs) and a bot software were also developed and incorporated into the prototype as part of the real time crowd intelligence for Twitter data gathering. The developed Twitter bot allows user to monitor every tweet being tweeted and can automate all or part of Twitter activity. Indeed, FAIS allows the user to query tweets from Twitter by a specific user and/or keyword using both Search and Streaming APIs. A Search API gathers historical geotag data while a Stream API monitors real-time geotagged tweets with shortlisting at-risk areas based on provided keywords. FAIS system can be used equally efficiently by stakeholders as a pervasive early warning system to take smart action during flood emergency situation.

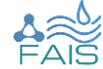

The prototype was tested for multiple hurricane events such as Hurricanes Dorian (2019), Lura (2020), and Ida (2021) across south and southeast US.

This manual introduces FAIS application and provides step by step instruction about how to run different modules for flood data analytics and assessment. FAIS prototype has several components integrated within the application that will be discussed in this manual including (1) USGS data collection, (2) SCDOT traffic image collection in real-time, (3) Twitter APIs and Geotagged data collection, (4) Data analytics, (5) Flood frequency analysis (FFA), and (6) Field data collection. **It should be noted that FAIS is a data engineering and analytics pipeline and doesn't deal with any flood simulation or forecasting tasks.**

## 2. System Requirements

To install FAIS python package (see here), users need to have an organizational installed Anaconda application. First install Django and FAIS packages by opening **Anaconda Application**. Select the "Environments" which allows user to create new development environment to install the packages.

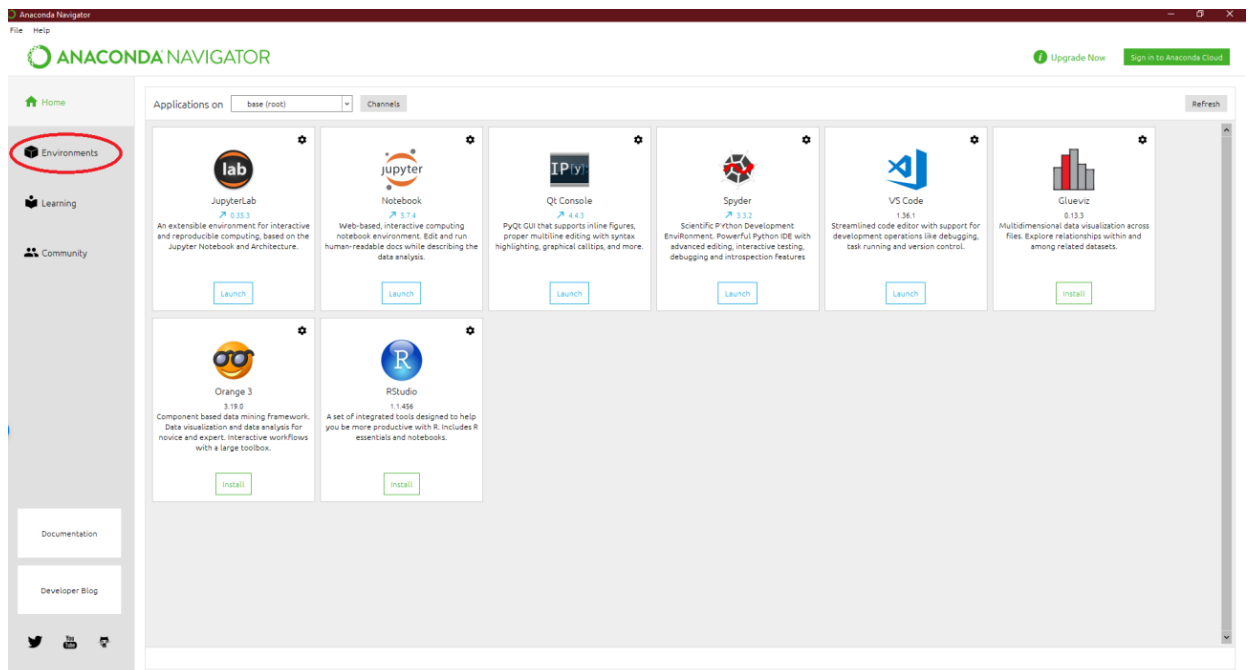

**Figure 1**. Organizational installed Anaconda application.

Now, select the "Create" button to create new Environment, now let's name it "FAIS" and select python 3.7 (or the newest version). After creating the environment, select the green play button next to your environment's name.

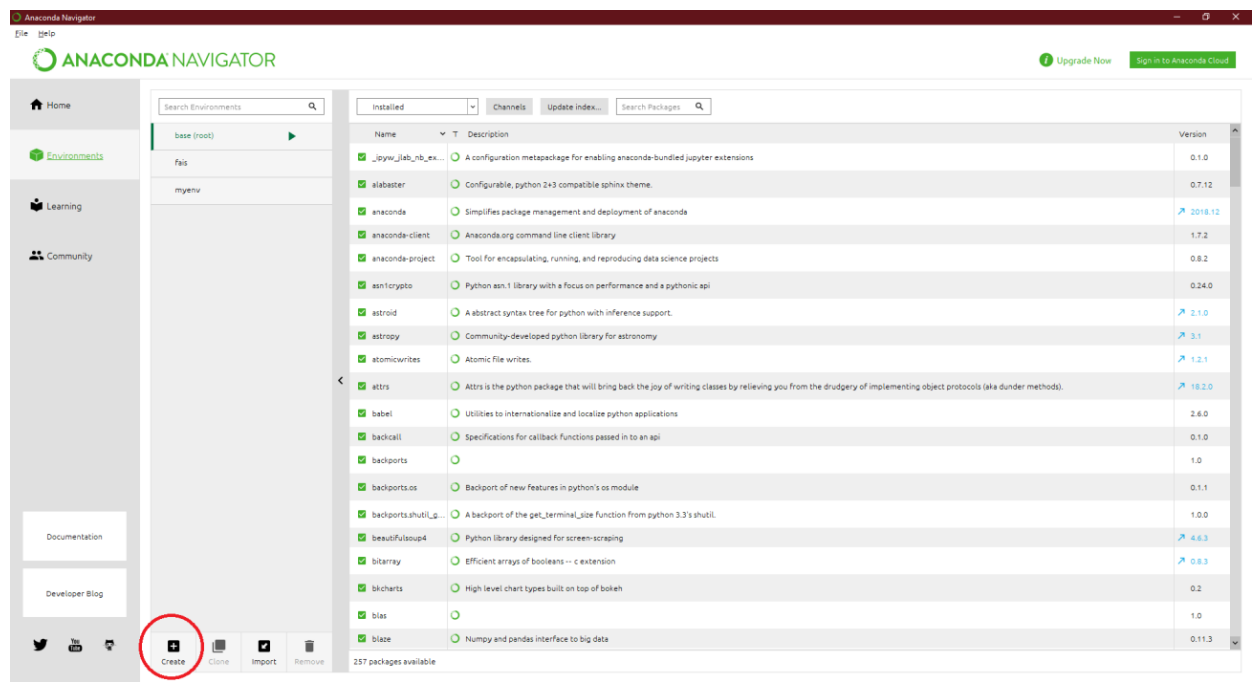

**Figure 2.** creating a new Environment for the FAIS installation.

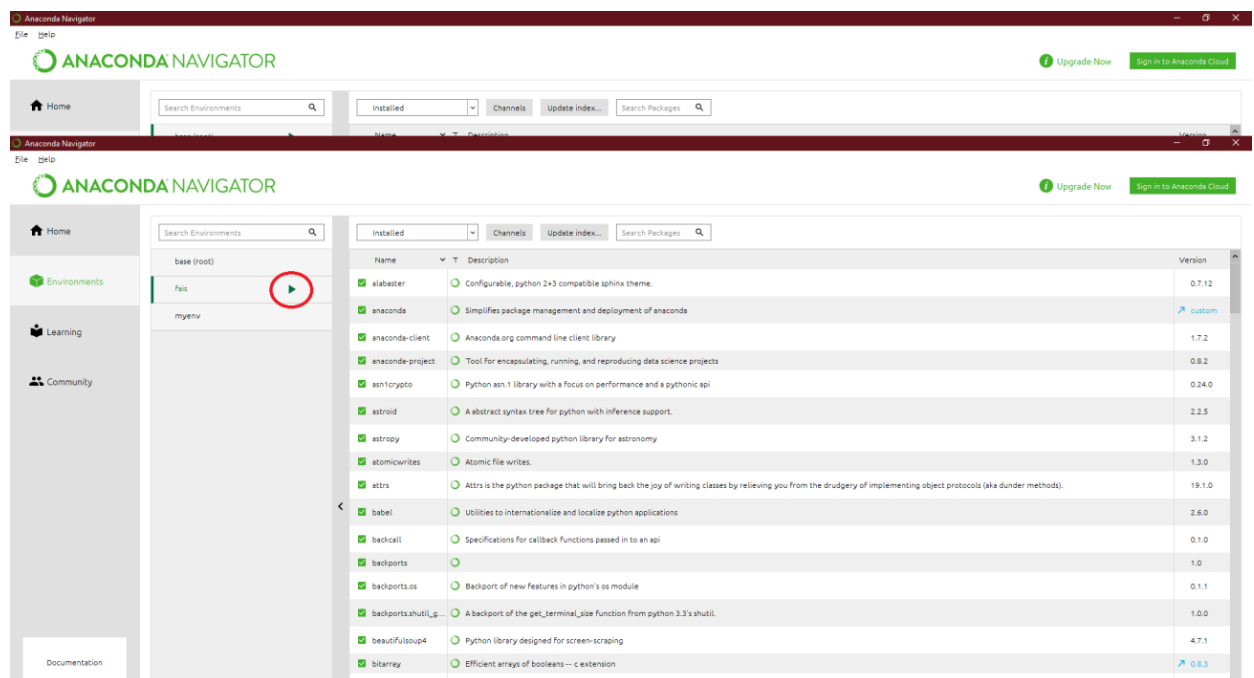

**Figure 3.** FAIS python package installed in a new environment.

This will create a dropdown, which you will select "Open Terminal". This option will open up a command prompt on your screen.

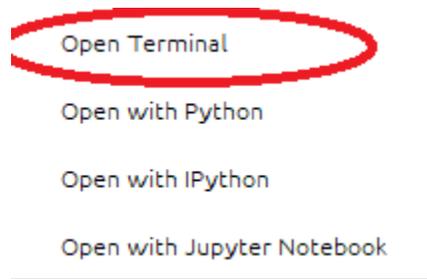

**Figure 4.** Launch FAIS using "Open Terminal"

Now, let's install some packages. First start with Django, please install Django package by typing pip install Django.

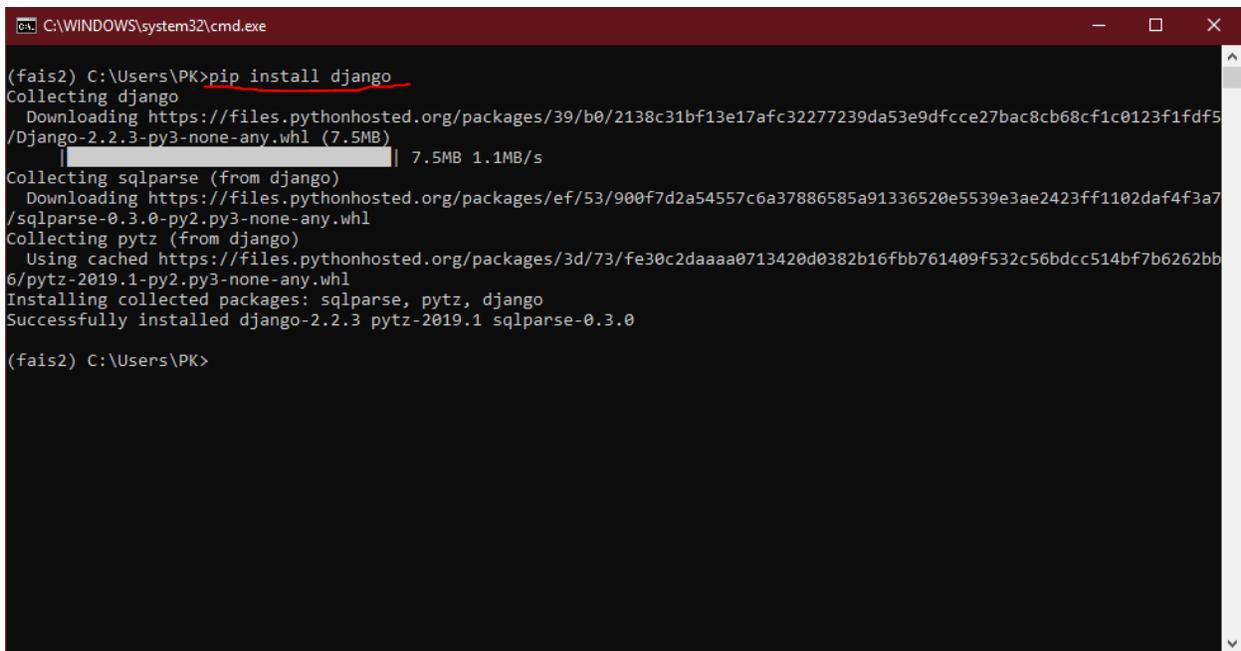

**Figure 5**. Command prompt to install the dependencies of the FAIS package.

Next, repeat the same method and install FAIS, and Django forms by typing

```
pip install fais
pip install django-form-utils
pip install --upgrade django-crispy-forms
pip install django-pandas
```

Note that this method could take some time. Congratulation, you finished the set-up process.

### 3. Start the Project

Use the provided source (see here) to download the project and place it in your document folder. On the terminal, navigate to the project folder, and start it by typing

```
cd Documents/Flood-Data-Analytics-App
python manage.py runserver
```

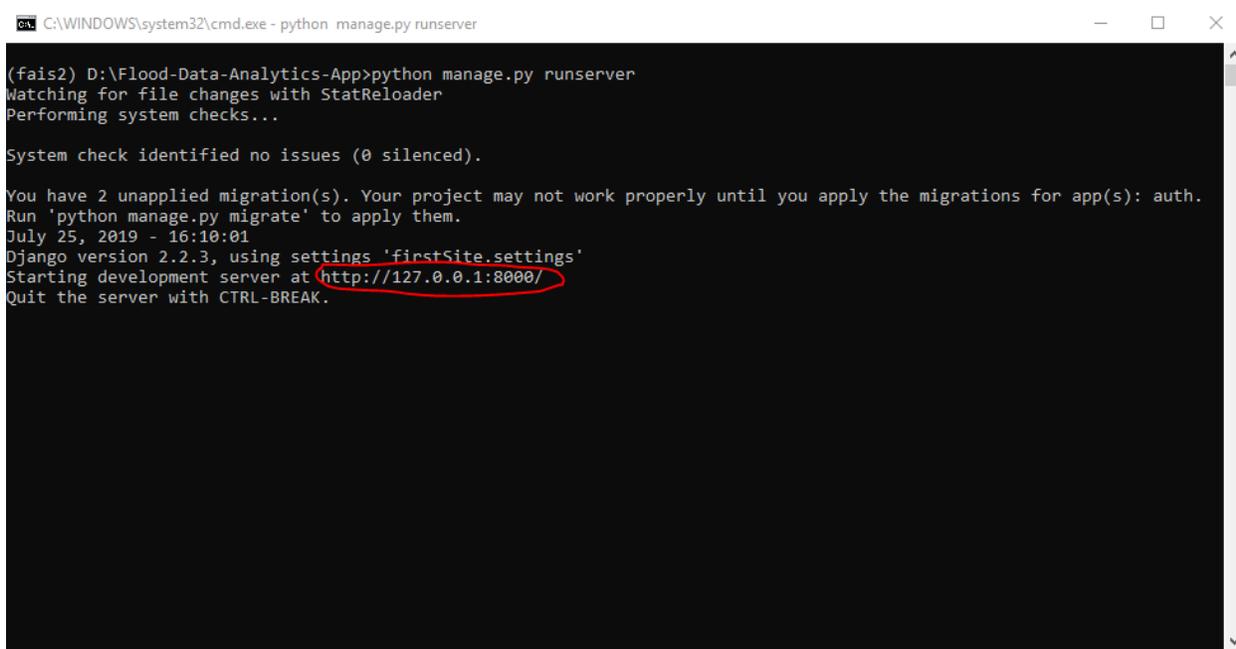

**Figure 6. How to extract FAIS server IP address.**

You should see responses like above on your terminal. Now on your browser, prefer Chrome, type 127.0.0.1:8000 or you can copy it from the terminal and paste it in your browser address bar, as well. You should see the print page of the application. Notice that there is a navigation bar on the right side of the application. This navigation bar allows you to move across different page of the application.

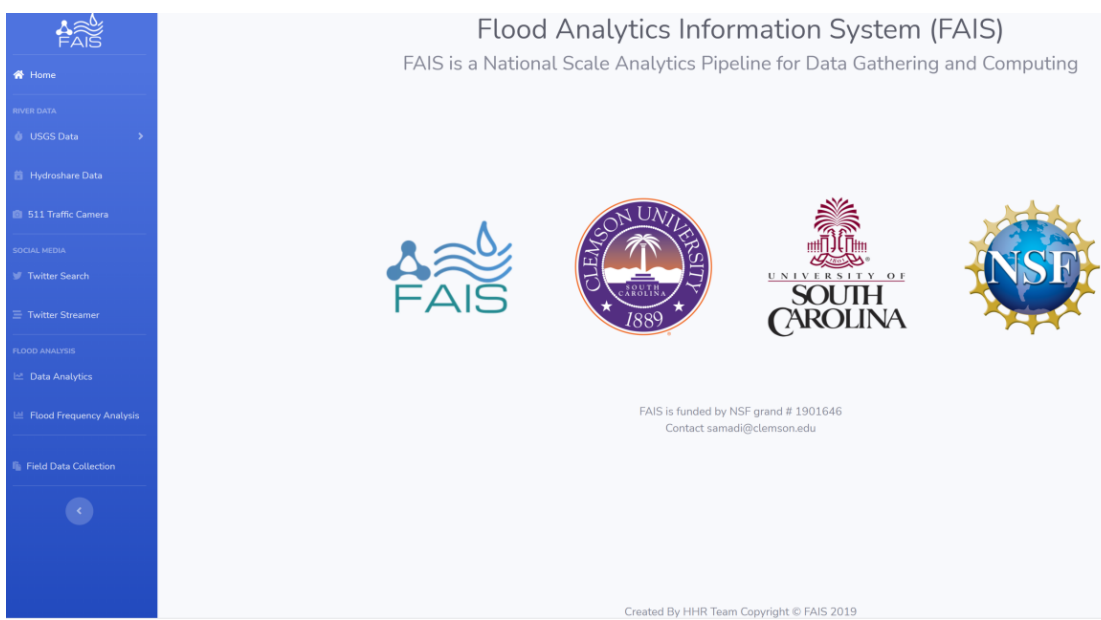

**Figure 7**. A print copy of the FAIS application.

The FAIS app can also be access online at https://floodanalytics.clemson.edu/. Chrome would be the best browser for working with a web FAIS.

### 4. USGS Real Time Data Collection

Click on USGS Data, there are three types of data to gather. Now click on Real-Time Flood Data.

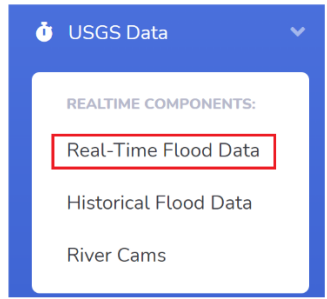

**Figure 8.** USGS data dropdown menu.

This page allows you to view the current flood reading in the selected State. The user can select the new State by opening the dropdown button. Now, select South Carolina (SC) to view the current flood reading. You can also find an URL associated with each station that refers you to the original data at the USGS website. You can also download the data as a .CSV format file by clicking on "Generate CVS Report".

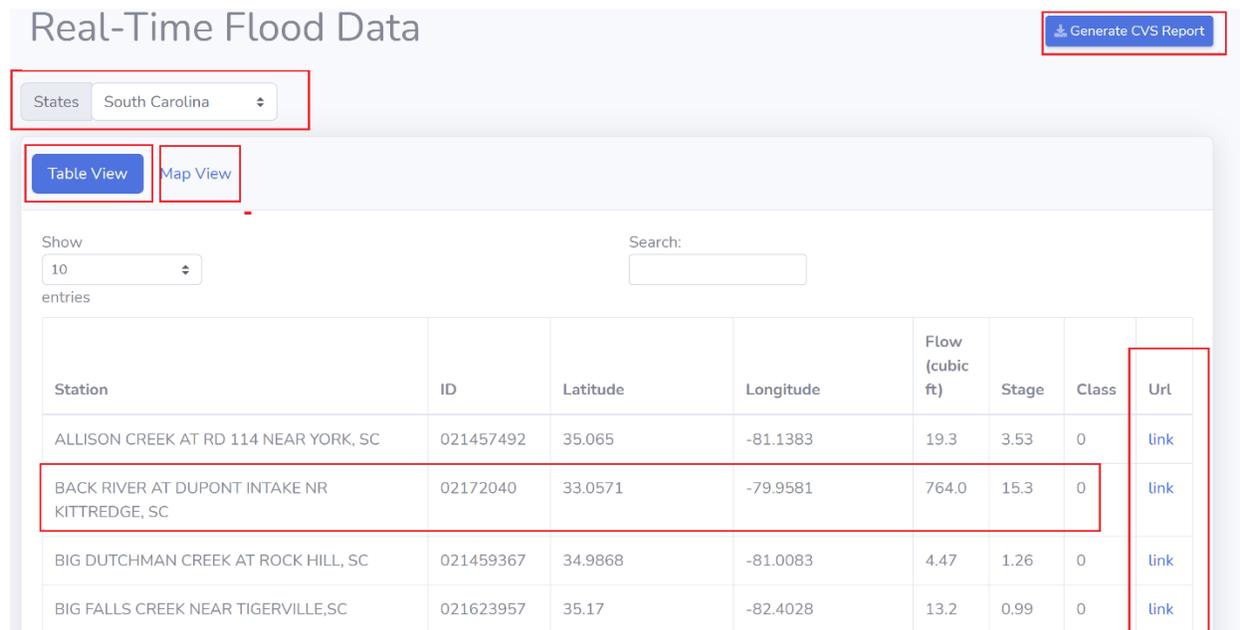

**Figure 9.** USGS real-time flood data gathering interface.

Select Map View option to see the geolocation of the USGS stations. Find Back River at DUPONT INTAKE NR KITTREDGE, SC on the map! and click on the station to see current flood reading, station id, and latitude and longitude of the gauging station.

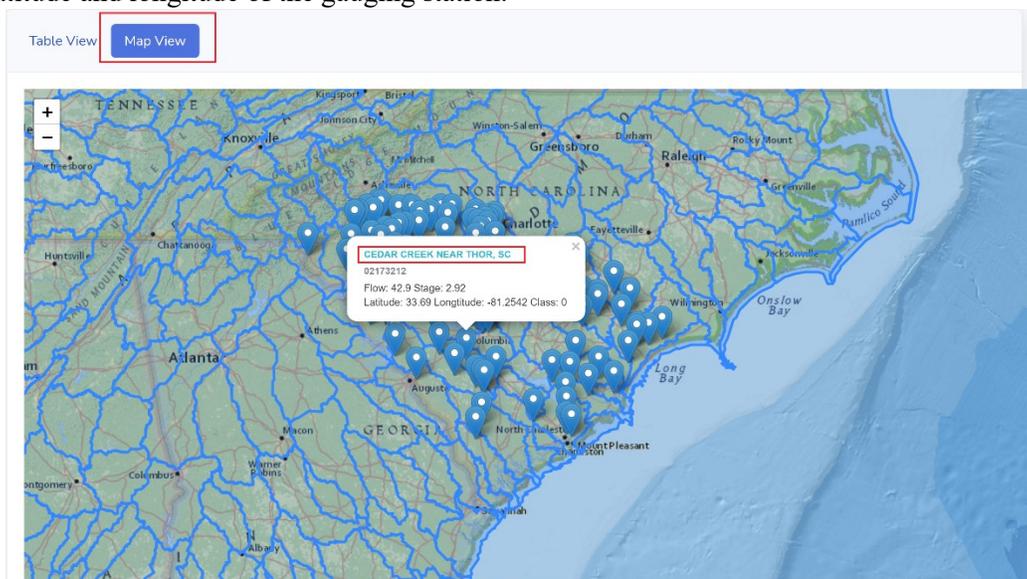

**Figure 10.** A map of the USGS real-time flood stations.

## 5. USGS Historical Data

In this section, we will be performing historical flood data gathering from USGS database. Navigate the Historical Flood Data to display the historical data page. To gather the data, user must first select the target state, which will then populate a list of available gauging stations to choose from. User can then select start and end date for the data. The result will show the datetime, discharge ($ft^3/s$), and gauge height (ft). Here you can see plots of your data during the time period that you selected. You can also download the data as a .CSV format file.

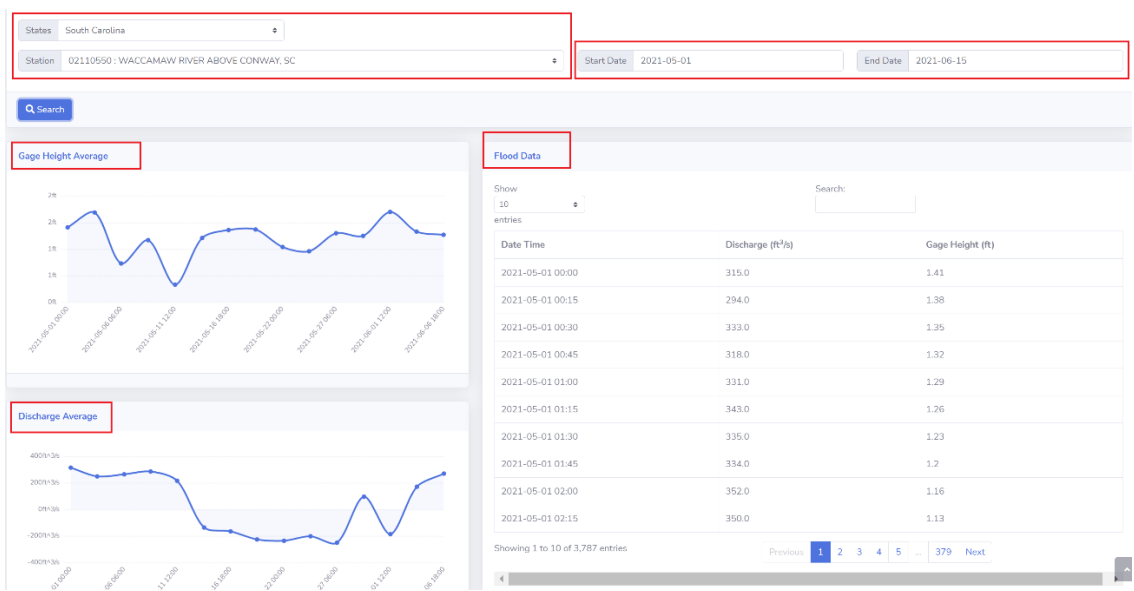

**Figure 11.** USGS historical flood data gathering interface.

## 6. USGS River Cams

In addition to flood gauging stations, USGS also established several river cameras to monitor the river system particularly during heavy storm events. You can select "River Cams" on the task bar to navigate to river images. This page allows you to view the current river camera stations and images. Note that this functionality is only available for some stations where the cams are active (if you receive an error that means the USGS portal stopped sending the response to the FAIS app). You can stream the images if your goal is to work on image data and understand flooding impacts on nearby locations/roads in real -time. To do so, just hit "Download Images" on the right corner and then save images in your desktop/local/cloud repository.

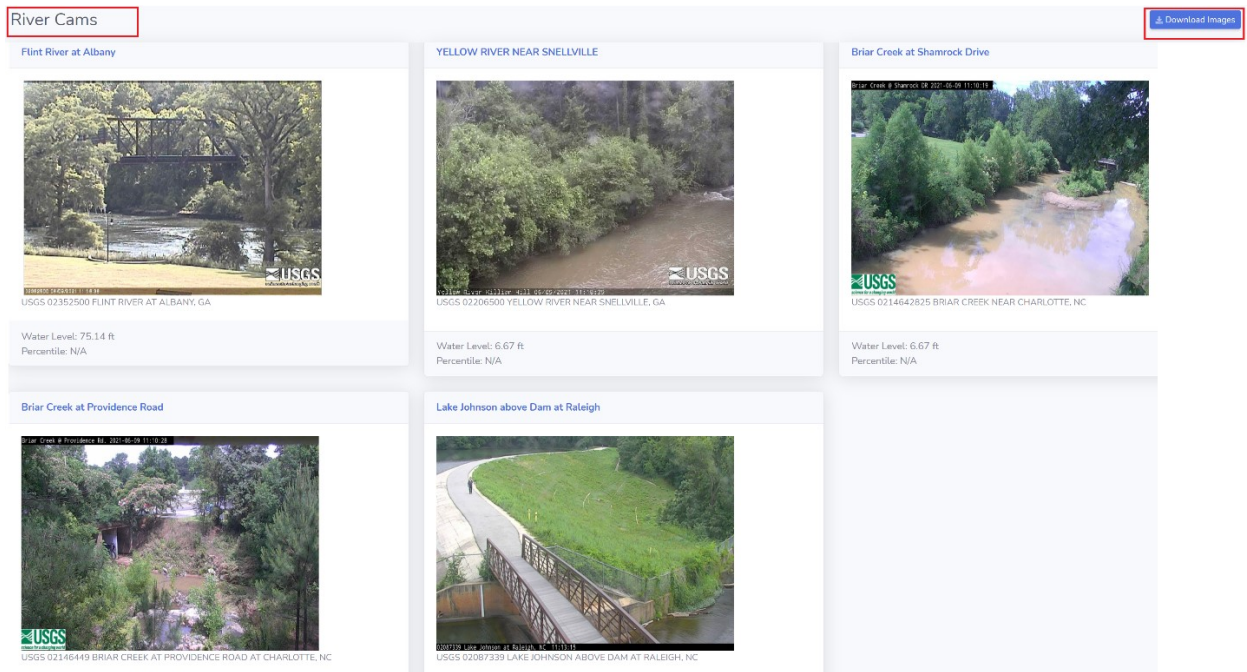

**Figure 12.** The USGS web cameras interface.

## 7. Twitter Data Gathering and Analysis

In this section, we will be performing data gathering from Twitter. Navigate to Twitter Search section, this will display the Twitter Data in a time frame that the user defines. To gather the data, the user must input the target user id, the specific keywords, or both. The user needs to input the beginning and ending dates for target tweet collection. Note that the user can gather tweets from past 7 days only. The Twitter gathering tool may need a little longer time to gather tweets. To begin searching, hit the search button and the result will be displayed as tweets, posted date, source link, media link if available, and the sentiment. The sentiment shows whether or not the tweet has positive tone (1), natural tone (0), or negative tone (-1). This is important to study societal impacts of flooding and how the residents responded to the flooding and damages. You can also download the tweets as a .CSV format file. The tool will then clean the tweets using several keywords such as "Floods", "Damage", "Flooded Road", etc. and then intersect the relevant tweets with watershed boundary and nearby USGS flood gauge. This will identify the potential locations at risk of flooding in real-time. Readers are referred to Donratanapat et al., (2020) for more information.

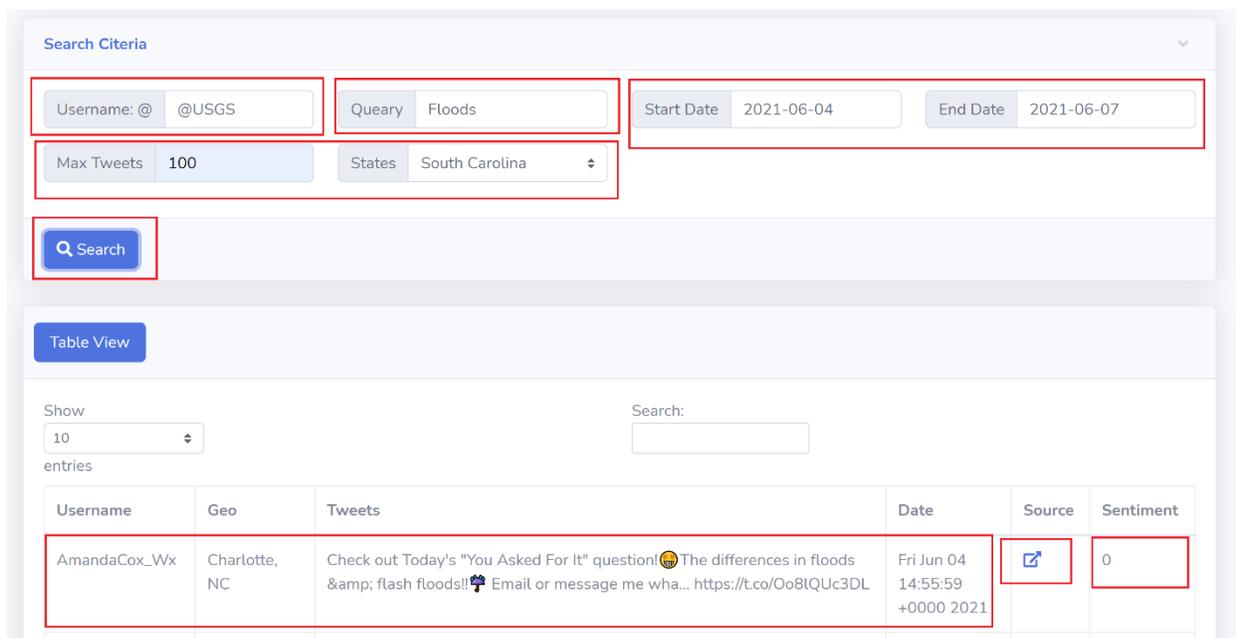

**Figure 13.** Streaming tweet interface.

### 8. Data Analytics

Convolutional neural networks (CNNs) are one form of deep learning algorithms widely used in computer vision which are used in the FAIS application to study flood images and assign learnable weights to various objects in the image. To develop CNN algorithms, we first built a training database service of >8000 images (image annotation service) including the image geolocation information by streaming relevant images from social media platforms, 511 traffic cameras, the US geological Survey (USGS) river cameras, and search engines videos. All these images were manually annotated to train the different models and detect a total of eight different object categories. We developed various CNNs architectures such as YOLOv3 (You look only once version 3; Redmon et al., 2016), Fast R-CNN (Region-based CNN; Girshick, 2015), Faster R-CNN (Ren etal., 2015), Mask R-CNN (He et al., 2017), SSD MobileNet (Single Shot MultiBox Detector MobileNet; Liu et al., 2016), and EfficientDet (efficient object detection ; Tan et al., 2020) to perform both object detection and segmentation simultaneously. Canny edge detection and aspect ratio concept are also included in the package for floodwater level estimation and classification. Based on the aspect ratio which is calculated by taking into consideration the area of the water surface detected within the image, water level is estimated. The water levels were then categorized into mild, moderate, and severe conditions to reflect flood severity and risk (see **Table 1**). The tool has the capability to train a large number of images and calculate floodwater levels and inundation areas which can be used to identify flood severity and risk. **Figure 14** displays the workflow of multiple CNN algorithms for image label detection and inundation calculation tasks.

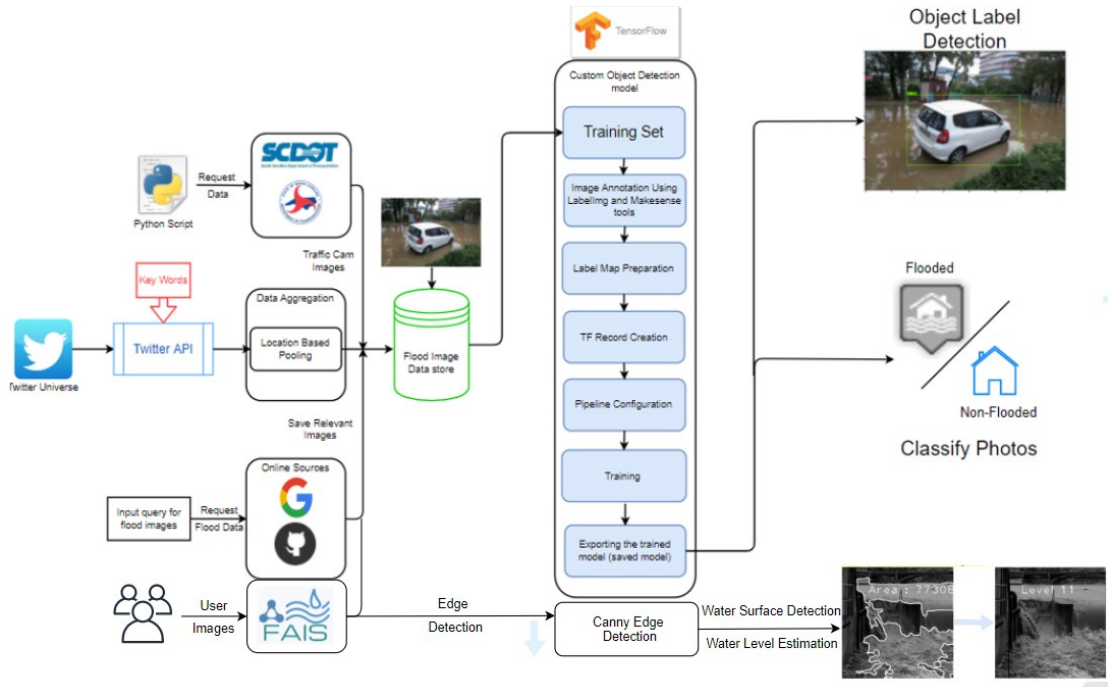

**Figure 14.** The workflow of data analytics module in the FAIS application.

**Table 1.** Water levels associated aspect ratios and flood severity and risk estimation.

| Water Level | Aspect Ratio | Flood Severity and Risk |
|---|---|---|
| Level 1 | >1.8 | Mild |
| Level 2 | 1.62 – 1.8 | |
| Level 3 | 1.44 – 1.62 | |
| Level 4 | 1.26 – 1.44 | |
| Level 5 | 1.08 – 1.26 | Moderate |
| Level 6 | 0.90 – 1.08 | |
| Level 7 | 0.72 – 0.90 | |
| Level 8 | 0.54 – 0.72 | |
| Level 9 | 0.36 – 0.54 | Severe |
| Level 10 | 0.18 – 0.36 | |
| Level 11 | <0.18 | |

Open "Data Analytics" option. Navigate your local folder or cloud to open an image. Once the image feeds to the module, it will then return the labels scores, flood depth, and the inundation areas.

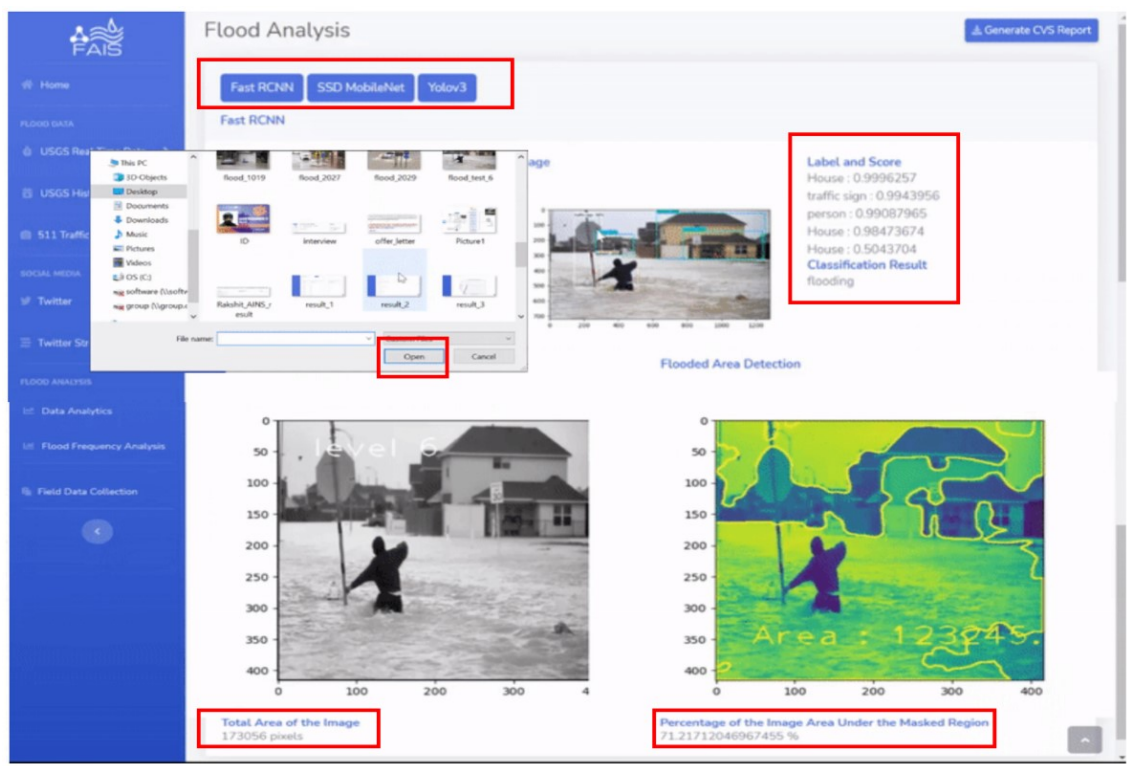

**Figure 15.** Data Analytics interface.

## 9. Flood Frequency Analysis

FAIS prototype also performs flood frequency analysis (FFA) to assist engineers in designing safe structures. The tool retrieves annual peak flow rates for provided years and calculates statistical information such as mean, standard deviation and skewness which are further used to create frequency distribution graphs. To perform flood frequency analysis for any station in US, you need to have the USGS (we selected USGS 02147500) station id number. Once you enter the number, hit the Search button, and then you will see the magic! FFA outcomes for different distribution along with the best fit will be displayed in the portal. You can also download the data as a .CSV format file.

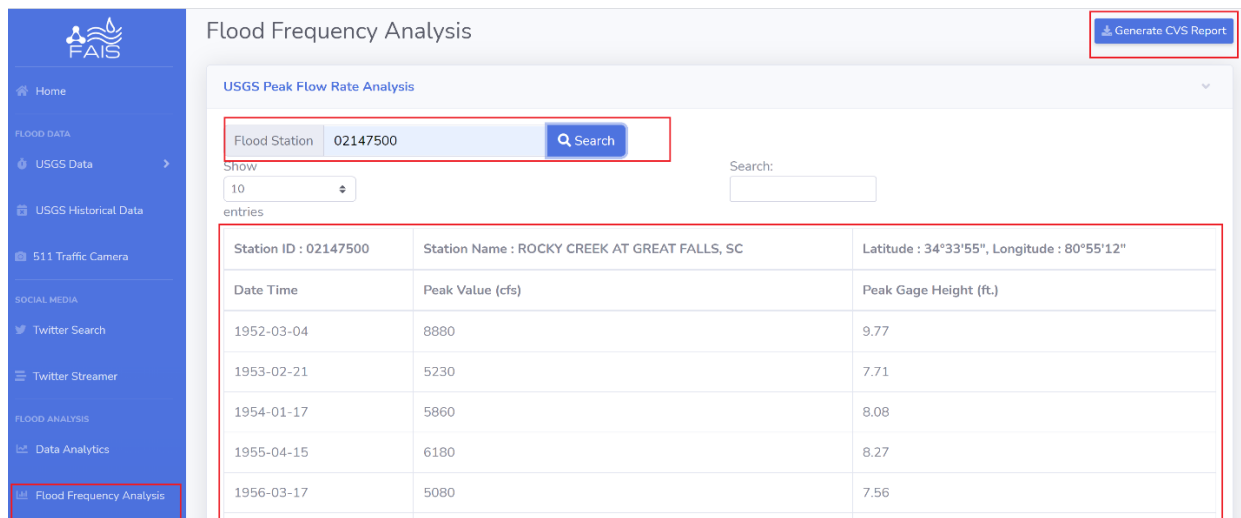

**Figure 16.** Flood Frequency Analysis interface.

You will then see several flood frequencies plots. If you are a professional Engineer or a decision maker, you may need to have the best fitted distribution to flood peak rates and FAIS app provides this analysis to user along with fitted distribution parameters. The tool fits various probability distributions including Normal, Lognormal, Gamma, Gumbel, Pearson Type III, etc. This analysis is useful in providing a measurement parameter to assess the damage corresponding to specific flow during flooding event. Along with civil infrastructure design, flood frequency analysis can be used in flood insurance and flood zoning activities. Accurate estimation of flood frequency not only helps engineers in designing safe infrastructure but also in protection against economic losses due to maintenance of structures. However , the accuracy of FFA estimates may vary using different probability distributions such as Pearson type III, Gamma, Normal, and other distributions. We recommend using Pearson Type III, but other distributions such as Gumbel function can be also used for a river system with less regulation and less significant reservoir operations, diversions, or urbanization effects. It is also important to quantify the precision of estimates, so the tool calculates flood frequencies within 95% confidence interval to provide the accuracy of the calculation. A 95% confidence interval is a range of values (upper and lower) that you can be 95% certain contains the true mean of the flood frequencies values/data population.

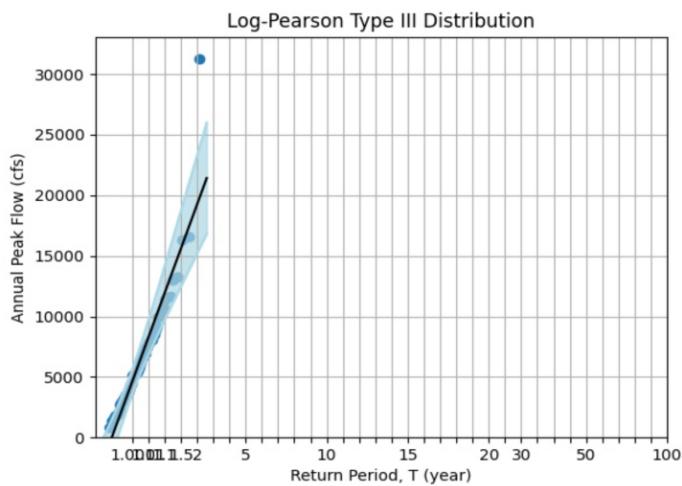

**Figure 17.** Log-Pearson Type III fitted plot to the USGS02147500 flood peak data.

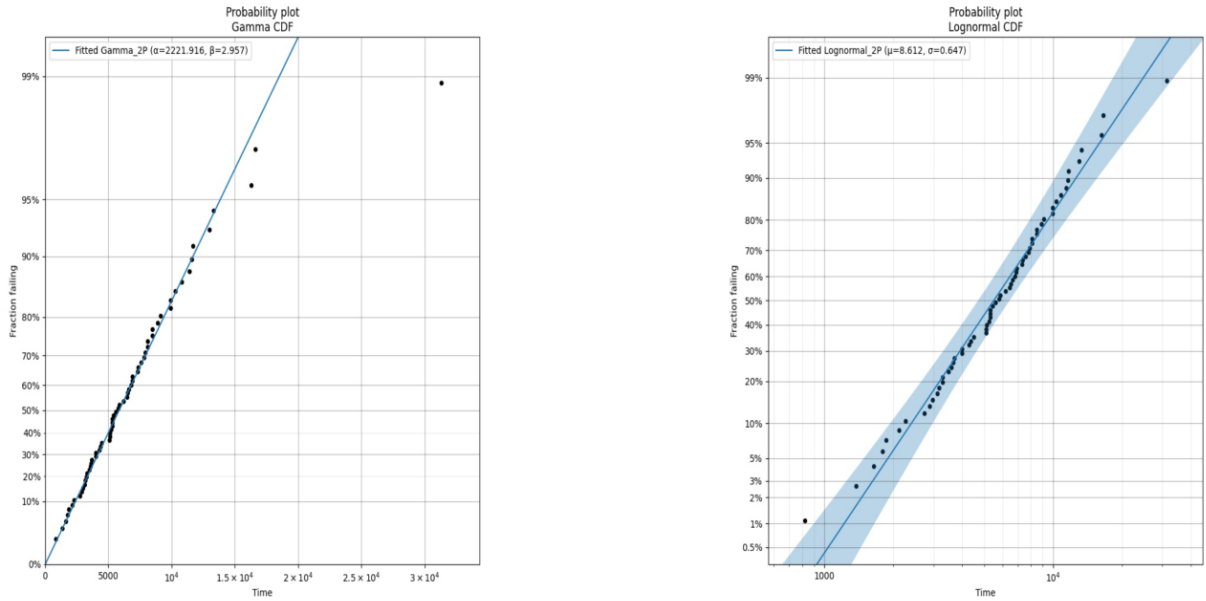

**Figure 18.** Fitted Gamma and Longnormal CDF plots for the USGS02147500 flood peak data.

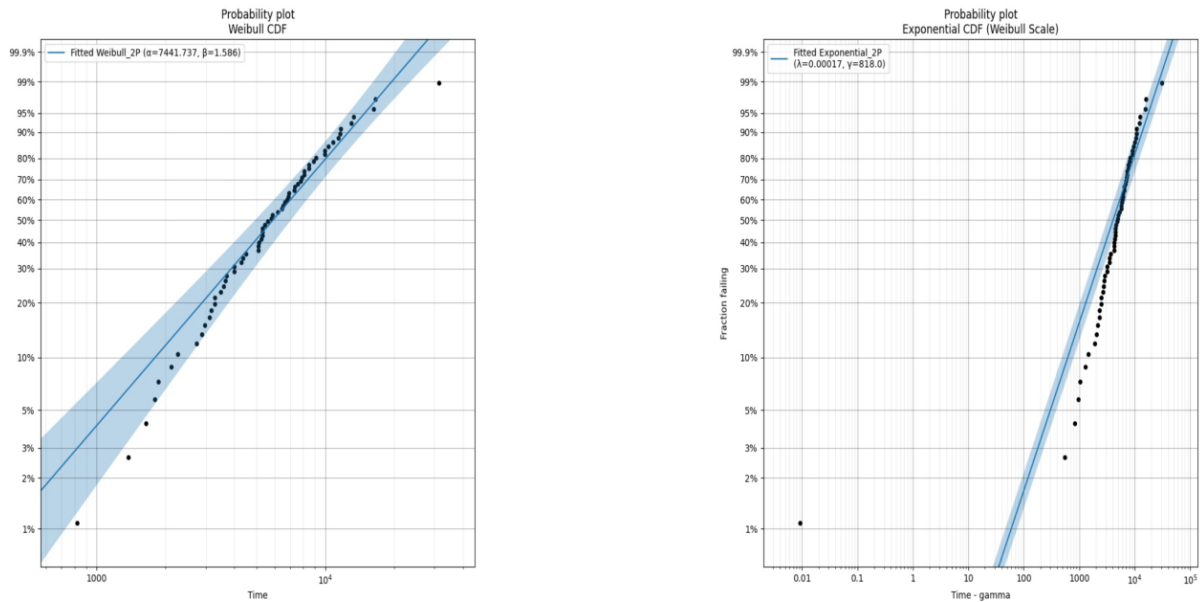

**Figure 19.** Fitted Weibull and Expomemtial CDF plots for the USGS02147500 flood peak data.

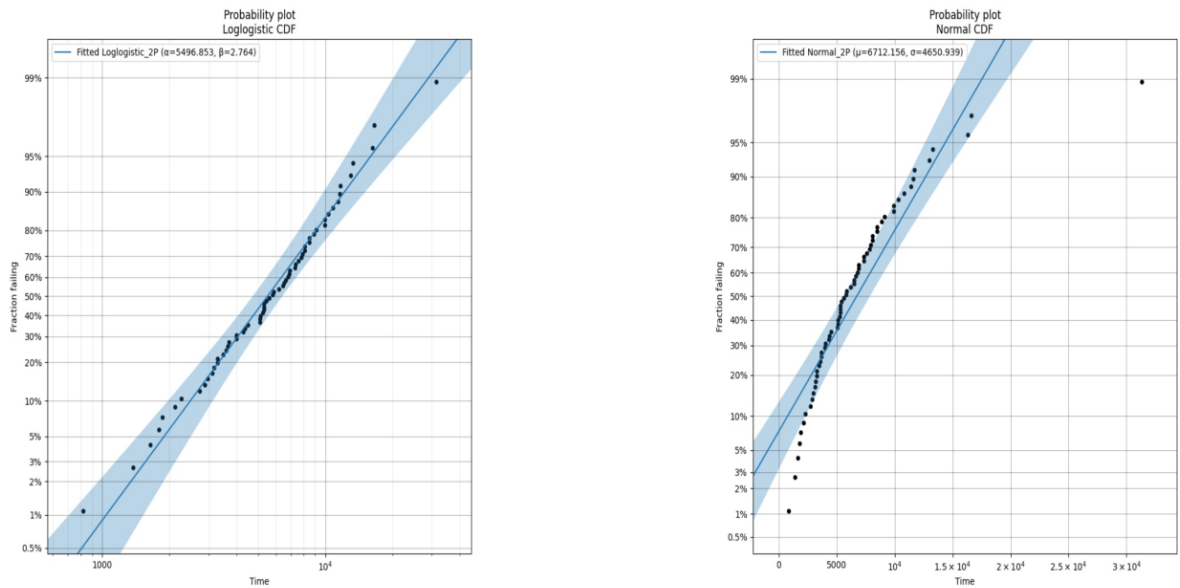

**Figure 20.** Fitted Loglogistic and Normal CDF plots for the USGS02147500 flood peak data.

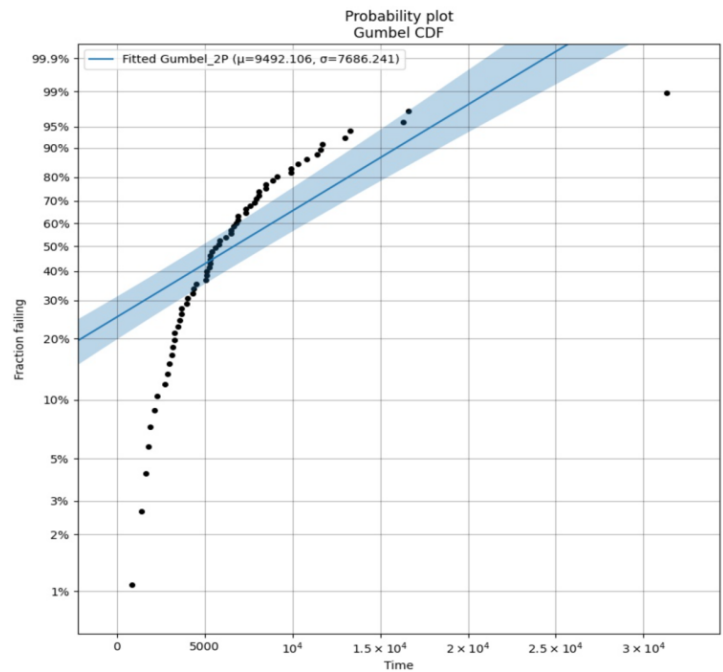

The best fitting distribution was Loglogistic_2P with which had parameters [5.49685308e+03 2.76406643e+00 0.00000000e+00]

**Figure 21.** Fitted Gumbel CDF plot for the USGS02147500 flood peak data.

## 10. Field Data Collection

This section enables user to report local flooding and collect and send the data to our team through FAIS application. The user can report the flood depth, location, etc. and submit the data to our team. This will help us to develop and validate our flood simulation models. Users have also the option to upload any images/video of the flood event that they observed!

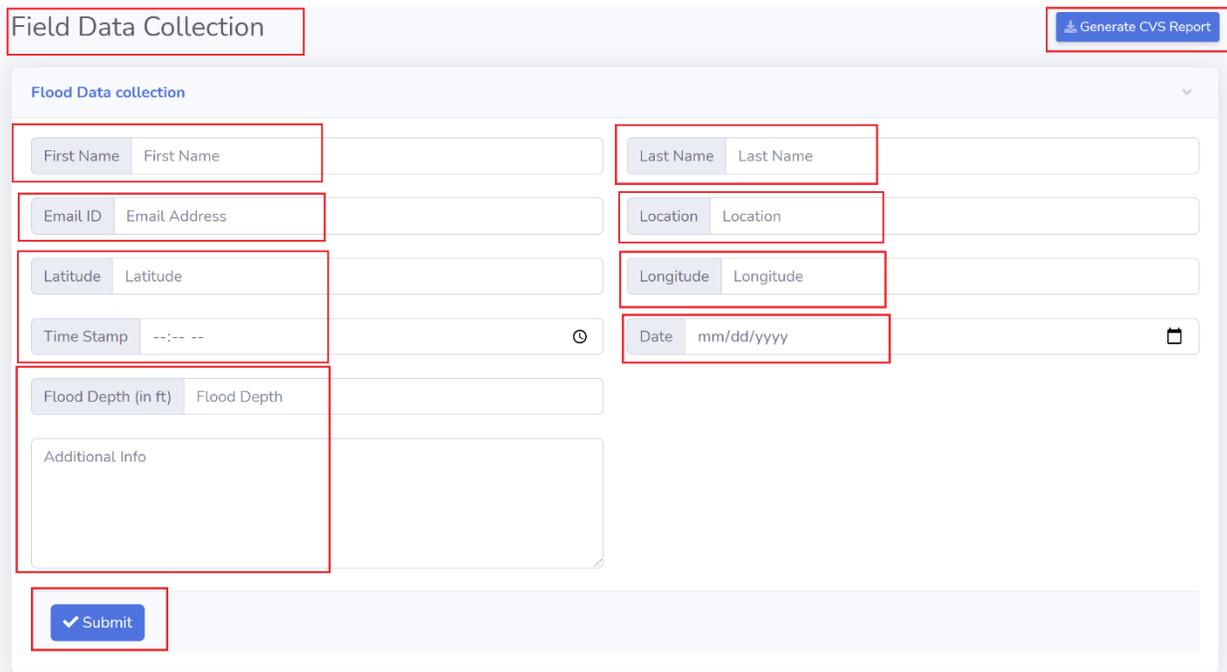

**Figure 22.** Flood data collection interface.

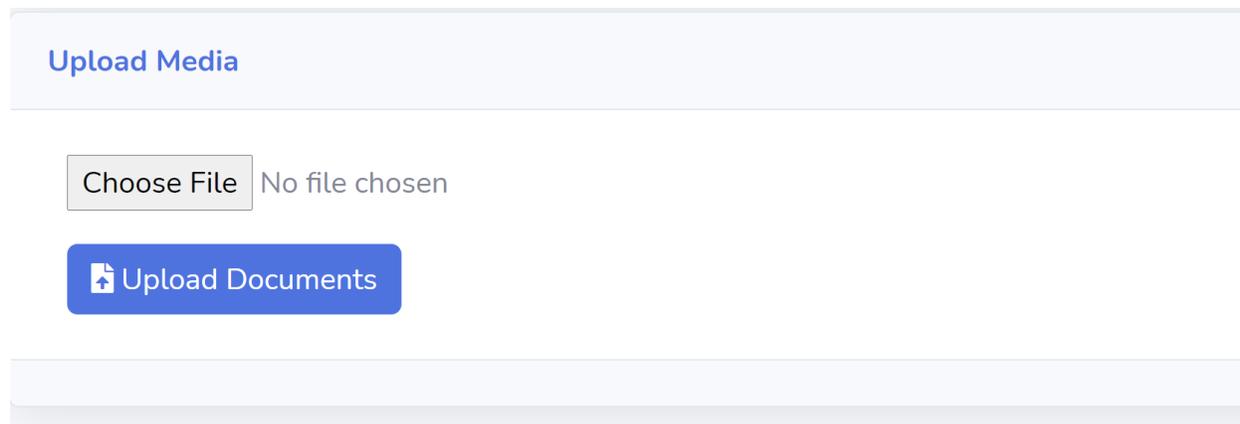

**Figure 23.** Flood data collection interface.

### 11. Software and Data Availability

FAIS-v4.00 application is publicly available at Clemson-IBM cloud service. FAIS python package is also available at Git and PIP . The data and flood images used to perform this research is available upon request.

### 12. Acknowledgements

This research is funded by the U.S. National Science Foundation (NSF) Directorate for Engineering under grant CBET 1901646. Any opinions, findings, and discussions expressed in this manual are those of the author and do not necessarily reflect the views of the NSF. The author also acknowledges IBM company for providing free credits to deploy and sustain FAIS application. The author also acknowledges the contribution of Hydrosystem and Hydroinformatics Research (HHR) students at Clemson University.

### 13. References

1. Donratanapat, N., Samadi S., Vidal, M.J., Sadeghi Tabas, S. 2020. A national-scale big data prototype for real-time flood emergency response and management. Environmental Modelling & Software. DOI: 10.1016/j.envsoft.2020.104828.
2. Girshick, R., 2015. Fast r-cnn. In Proceedings of the IEEE international conference on computer vision https://doi.org/10.1109/ICCV.2015.169.
3. He, K., Gkioxari, G., Dollár, P. and Girshick, R., 2017. Mask r-cnn. In Proceedings of the IEEE international conference on computer vision. DOI: 10.1109/ICCV.2017.322.
4. Redmon, J., Divvala, S., Girshick, R. and Farhadi, A., 2016. You only look once: Unified, real-time object detection. In Proceedings of the IEEE conference on computer vision and pattern recognition. DOI: 10.1109/CVPR.2016.9.
5. Ren, S., He, K., Girshick, R., and Sun, J. 2015. "Faster r-cnn: towards real-time object detection with region proposal networks," in Advances in Neural Information Processing Systems (Montreal, QC), 91–99.
6. Tan M, Pang R, Le QV (2020) Efficientdet: scalable and efficient object detection. In: Proceedings of the IEEE/CVF Conference on Computer Vision and Pattern Recognition, pp 10781–10790
7. Wei Liu, Dragomir Anguelov, Dumitru Erhan, Christian Szegedy, Scott Reed, Cheng-Yang Fu: "SSD: Single Shot MultiBox Detector", 2016; arXiv:1512.02325.